\documentclass[twoside,journal]{IEEEtran}
\normalsize
\usepackage{cite}
\usepackage{algorithm}
\usepackage{algpseudocode}
\usepackage{amsmath,amssymb,amsfonts,amsthm}
\usepackage{graphicx}
\usepackage{textcomp}
\usepackage{xcolor}
\usepackage{subfig}
\usepackage{bm}

\def\BibTeX{{\rm B\kern-.05em{\sc i\kern-.025em b}\kern-.08em
    T\kern-.1667em\lower.7ex\hbox{E}\kern-.125emX}}

\newtheorem{remark}{Remark}

\makeatletter
\renewcommand\normalsize{%
\@setfontsize\normalsize\@xpt\@xiipt
\abovedisplayskip 5\p@ \@plus3\p@ \@minus3\p@
\abovedisplayshortskip \z@ \@plus3\p@
\belowdisplayshortskip 5\p@ \@plus3\p@ \@minus3\p@
\belowdisplayskip \abovedisplayskip
\let\@listi\@listI}
\makeatother
\begin{document}

\title{Decision Transformer for UAV-Mounted RIS-Assisted Dynamic D2D Communications}
\author{
        Yaxuan Liu
%
\thanks{Y. Liu is with the School of Computer Engineering, School of Artificial Intelligence, Jiangsu Second Normal University, Nanjing 210013, China (e-mail: Carol\_liuyx@163.com).}
%
%
}

\maketitle

\begin{abstract}
This paper studies unmanned aerial vehicle (UAV)-mouted reconfigurable intelligent surface (RIS)-assisted device-to-device (D2D) communication with stochastic link activation. It models UAV motion and attitude, time-varying Rician angles, and angle-dependent RIS reflection. A joint optimization of UAV trajectory, attitude, and RIS phases is formulated to maximize average sum rate under mobility, energy, and hardware constraints. The problem is addressed using deep reinforcement learning and a Decision Transformer trained on expert trajectories from multiple scenarios. Results demonstrate effective cross-scenario generalization, with zero-shot transfer outperforming direct DRL transfer and online fine-tuning achieving competitive performance with fewer interactions.

\end{abstract}


\section{Introduction}
Reconfigurable intelligent surface (RIS) has emerged as a promising low-cost and energy-efficient technology for enhancing wireless signal coverage and communication quality \cite{RIS}. Conventionally, RIS devices are fixedly deployed on walls or stationary infrastructures, which severely limits their flexibility and adaptability in dynamic communication scenarios such as vehicular networks and industrial wireless systems. Such fixed deployment fails to cope with time-varying channel conditions and frequently occurring link blockage, resulting in degraded beamforming performance and unreliable transmission. To address these limitations, the integration of RIS with unmanned aerial vehicles (UAVs) provides an effective solution. Benefiting from the high mobility and flexible maneuverability of UAVs, the UAV-mounted RIS system can dynamically adjust its spatial position and attitude in real time. Different from static RIS, this integrated framework can actively avoids wireless link blockage to maintain stable line-of-sight (LoS) propagation paths.

In \cite{1-1,1-2}, the authors jointly optimize UAV trajectory and RIS phase shifts to enhance the sum rate, physical-layer security and energy efficiency. However, These studies either assumed that the RIS is deployed on a stationary planar surface, or considered UAV-mounted RIS systems while ignoring the three-dimensional rotational attitudes of UAV. Several other works incorporate the three-dimensional rotational angles of UAVs as random jitter disturbances, without treating RIS orientation as an adjustable optimization variable and still constraining the RIS to a fixed plane. Several studies verified from an electromagnetic perspective that the reflection coefficients of RIS elements are dependent on incident angles \cite{EM1,EM2}. Though \cite{3-1,3-2} integrated this angle-variant reflection property into UAV-mounted RIS modeling, \cite{3-2} focused on the statistic scenarios, while \cite{3-1} relied on an oversimplified uniform linear array (ULA) steering model and neglected the intricate coupling among UAV pose, time-varying Rician channels, and element-wise local incident angles, which is critical for indoor propagation environments.

Notably, since the optimization problem in UAV assisted communications are non-convex and coupling, deep reinforcement learning (DRL) algorithms are always adopted to resolve theses complicated problems. For example, deep deterministic policy gradient (DDPG) algorithm was used in \cite{1-1} to jointly optimize the UAV trajectories and RIS phase shift. Authors in \cite{SAC-review} employed soft actor-critic (SAC) to jointly optimize UAV trajectory and resource allocation for maximizing computation bits under a fairness constraint. However, conventional DRL policies often lack robustness to changes in communication environments, with performance degrading in unseen scenarios and adaptation requiring costly online interactions \cite{zhangjie}. The Decision Transformer (DT) addresses this issue by framing reinforcement learning as conditional sequence modeling \cite{DT}. Based on the Transformer architecture, it predicts actions from returns-to-go, states, and interaction history, enabling efficient policy learning from offline trajectories without value estimation or policy-gradient updates.

Motivated by these, this work studies a dynamic indoor communication system assisted by a UAV-mounted RIS configured as a uniform planar array (UPA), in which communication links between users are stochastically established over time. The UAV trajectory, three-dimensional attitude, and RIS phase shifts are jointly optimized while accounting for the time-varying arrival and departure angles of the Rician LoS components and the incident-angle-dependent responses of the RIS elements. A Decision Transformer is pre-trained offline on high-quality DRL trajectories across multiple scenarios, enabling zero-shot control and efficient online adaptation in unseen deployments with fewer interactions.

\section{System Model}

We focus on the dense industrial manufacturing scenario where exist $K$ single-antenna device-to-device (D2D) communication devices, denoted by the set $\mathcal{D} = \left\{ {D{U_1},D{U_2},...,D{U_K}} \right\}$. Pairwise communication is established among these D2D devices by reusing spectrum resources on licensed frequency bands. Within each communication time interval, at most $\lfloor K/2 \rfloor$ D2D pairs are randomly activated with random user pairing. A UAV-mounted RIS is deployed to assist communications, where both desired signals and interference propagate via RIS reflection.

We consider that all D2D devices possess different altitudes, and the UAV supports flexible altitude adjustment rather than hovering at a constant height. For this practical scenario, we adopt an incident-angle-dependent RIS reflection coefficient model. To accurately characterize the spatial incident angles, the RIS is equipped with a UPA. In contrast to conventional RIS models with ideal unitary reflection coefficients, the vertical height differences among D2D devices lead to dynamic variations in the incident angle impinging on the RIS, which further alters the resultant reflection coefficients. Accordingly, it is essential to dynamically adjust the three-dimensional attitude angles of the RIS-mounted UAV to deliberately regulate the RIS incident angle. Based on this physical mechanism, we jointly optimize the UAV flight trajectory, three-dimensional attitude angles, and RIS phase shifts to enhance the overall communication performance.
\begin{figure}
  \centering
  \includegraphics[width=8.3cm,height=4.8cm]{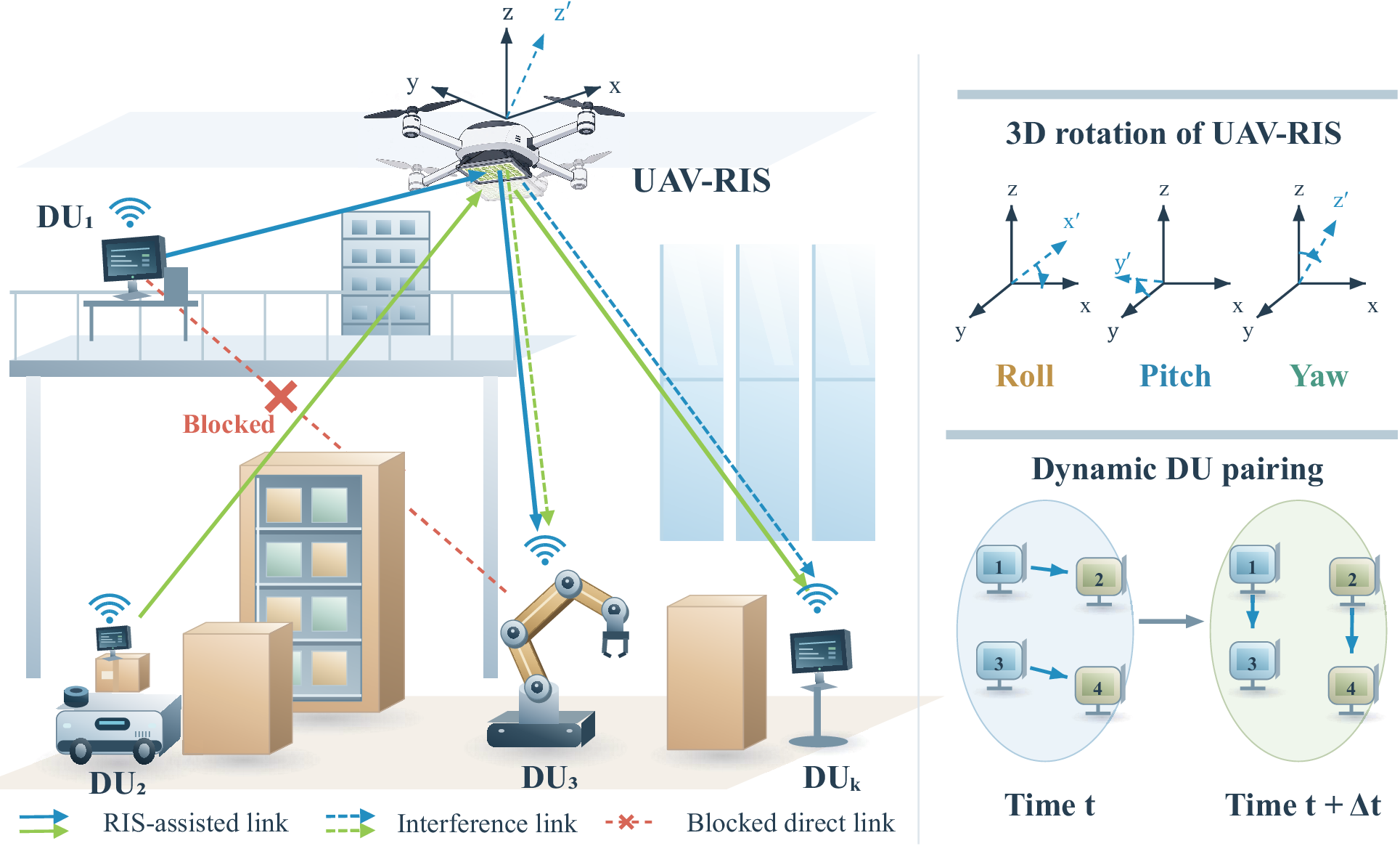}
  \caption{UAV-Mounted RIS-Assisted Dynamic D2D Communication System.}\label{fig:systemmodel}
\end{figure}
\subsection{Channel Model}
Mutual interference emerges when multiple D2D pairs transmit concurrently within the same time slot. Severe ground blockages are assumed in this work, such that the direct paths between D2D users can be neglected. Since the UAV operates at altitude, LoS links are maintained between ground users and the RIS. Accordingly, the channel between D2D user $ {D{U_k}}$ and the RIS is modeled by Rician fading, which yields
\begin{align}\label{eq:Rician}
{\mathbf{h}_{k,R}} = \sqrt {\frac{\kappa }{{\kappa  + 1}}} {\mathbf{\bar h}_{k,R}} + \sqrt {\frac{1}{{\kappa  + 1}}} {\mathbf{\tilde h}_{k,R}},
\end{align}
where $\kappa$ denotes the Rician factor, ${\mathbf{\bar h}_{k,R}} = \sqrt {d_{k,R}^{ - {\alpha _L}}} {\mathbf{\bar g}_{k,R}}$ stands for the LoS component where $d_{k,R}$ is the distance between $ {D{U_k}}$ and RIS, ${\alpha _L}$ represents the path-loss exponent for the LoS path, and ${\bar g_{k,R}}$ denotes the receiving array response of the LoS component. Since UPA is considered for RIS, we denote the numbers of elements along the x-axis and y-axis as $N_x$ and $N_v$, respectively, yielding a total of $N=N_x\times N_y$ units. The angle of arrival (AoA) at the RIS is $\left( {{\phi _{k,R}},{\vartheta _{k,R}}} \right)$, where $ {\phi _{k,R}}$ and ${\vartheta _{k,R}} $ represent the azimuth and elevation angles of the incident electromagnetic wave with respect to the RIS local coordinate system. Therefore, the receiving array response at the UAV-mounted RIS is expressed by ${\mathbf{\bar g}_{k,R}} = {{\mathbf{a}}_y}\left( {{\phi _{k,R}},{\vartheta _{k,R}}} \right) \otimes {{\mathbf{a}}_x}\left( {{\phi _{k,R}},{\vartheta _{k,R}}} \right)$
where
\begin{align}
  {{\mathbf{a}}_x}\left( {{\phi _{k,R}},{\vartheta _{k,R}}} \right) = {\left[ {1, \ldots ,{e^{ - j\frac{{2\pi }}{\lambda }d({N_x} - 1){u_{k,R}}}}} \right]^T},
\end{align}
and
\begin{align}
  {{\mathbf{a}}_y}\left( {{\phi _{k,R}},{\vartheta _{k,R}}} \right) = {\left[ {1, \ldots ,{e^{ - j\frac{{2\pi }}{\lambda }d({N_y} - 1){v_{k,R}}}}} \right]^T},
\end{align}
where ${u_{k,R}} = \sin {\vartheta _{k,R}}\cos {\phi _{k,R}}$ and ${v_{k,R}} = \sin {\vartheta _{k,R}}\sin {\phi _{k,R}}$ represent the spatial frequencies along the local $x$-axis and $y$-axis, respectively, $d$ denotes the inter-element spacing, $\lambda$ denotes the carrier wavelength and $\otimes $ is the Kronecker product operation. Similarly, the transmit array response, defined by its angle of departure (AoD), can be calculated and denoted as ${{\mathbf{\tilde g}}_{R,k}}$. In addition, ${\mathbf{\tilde h}_{k,R}} = \sqrt {d_{k,R}^{ - {\alpha _L}}} {\mathbf{\tilde g}_{k,R}}$ denotes the non-line-of-sight (NLoS) component where ${\mathbf{\tilde g}_{k,R}}$ follows a circularly symmetric complex Gaussian distribution with zero mean and unit variance. Similarly, the channel from the RIS to $ {D{U_i}}$ is denoted as ${\mathbf{h}_{R,i}}$, which possesses the identical mathematical structure as \eqref{eq:Rician}.

\subsection{Angle Dependent RIS Reflection Model}
Specially, in this paper we adopts the practical reflection coefficient modeling framework for RIS proposed in \cite{Unified}. We first define $\mathbf{\Theta}$ as the phase-shift matrix of the RIS, which is formulated as
\begin{align}
 \mathbf{\Theta} = \text{diag}\left\{ {\left( {r_1^{pre},...,r_N^{pre}} \right)} \right\},
\end{align}
where $r_n^{pre} = \left| {r_n^{pre}} \right|{e^{j\angle r_n^{pre}}}\left( {n = 1,...,N} \right)$ denotes the preset complex reflection coefficient. According to \cite{Unified}, the practical reflection coefficient depending on the incident angle is obtained as
\begin{align}\label{eq:practical r}
  {r_n^{pra}}\left( {{\theta _{in}}} \right) = \frac{{ - \eta {\lambda _1}}}{{2\cos {\theta _{in}} + \eta {\lambda _1}}} + \frac{{{\lambda _2}\cos {\theta _{in}}}}{{{\lambda _2}\cos {\theta _{in}} + 2\eta }},
\end{align}
where $\theta _{in}$ is the electromagnetic wave incident angle, and
\begin{align}
  {\lambda _1} = \frac{2}{\eta }\frac{{1 - {r_n^{pre}}}}{{1 + {r_n^{pre}}}},{\lambda _2} = \frac{2}{\eta }\frac{{1 + {r_n^{pre}}}}{{1 - {r_n^{pre}}}},
\end{align}
with the free-space wave impedance $\eta  = 120\pi \Omega$. We denote ${{\mathbf{\Theta }}^{pra}} = {\text{diag}}\left\{ {\left( {r_1^{pra},...,r_N^{pra}} \right)} \right\}$ as the practical phase matrix of RIS. This model characterizes how the preset reflection amplitude and phase of RIS elements vary with the incident angle of electromagnetic waves, and outputs the actual reflection coefficient under specific propagation scenarios. The calculation method of ${{\theta _{in}}}$ will be given in \textbf{Remark \ref{remark:angle caculation}}.

\subsection{Signal Model}
Based on the channel model and the RIS reflection model, the received signal at user $ {D{U_i}}$ can be formulated as
\begin{align}
  {y_i} = &\sqrt P \left( {{\mathbf{h}}_{k,R}^H{\mathbf{\Theta}^{pra} \mathbf{h}}_{R,i}^{}} \right){x_k} + \nonumber\\
  &\sum\limits_{j = 1,j \ne k}^K {{\beta _j}\sqrt P \left( {{\mathbf{h}}_{j,R}^H{\mathbf{\Theta}^{pra} }{h_{R,i}}} \right){x_j}}  + {n_i}
\end{align}
where $P$ denotes the transmit power of each D2D user, and we assume identical transmit power for all users in this paper, ${\beta _j \in \left\{ {0,1} \right\}}$ is the control coefficient, which indicates whether $ {D{U_j}}$ performs transmission in the current time slot. $x_j$ and $x_k$ represent the transmitted signals of $ {D{U_j}}$ and $ {D{U_k}}$, respectively; $n_i$ denotes the additive white Gaussian noise (AWGN) at receiver $ {D{U_i}}$. Accordingly, the achievable rate at $ {D{U_i}}$ can be formulated as
\begin{align}
  {R_i} = {\log _2}\left( {1 + \frac{{P\left( {{\mathbf{h}}_{k,R}^H{\mathbf{\Theta h}}_{R,i}^{}} \right)}}{{\sum\nolimits_{j = 1,j \ne k}^K {{\beta _j}P\left( {{\mathbf{h}}_{j,R}^H{\mathbf{\Theta }}{{\mathbf{h}}_{R,i}}} \right)}  + {\sigma ^2}}}} \right),
\end{align}
where ${\sigma ^2}$ is the noise power.

\subsection{UAV Motion Model}
Let the position coordinate of the UAV at time $t$ be ${{\mathbf{q}}_U}\left( t \right) = \left( {{x_U}\left( t \right),{y_U}\left( t \right),{z_U}\left( t \right)} \right)$, and its velocity and acceleration vectors along the $X$, $Y$, $Z$ axes be denoted as ${\mathbf{v}_U} = \left[ {{v_x},{v_y},{v_z}} \right]$ and ${\mathbf{a}_U} = \left[ {{a_x},{a_y},{a_z}} \right]$, respectively. After a time slot interval $\Delta t$, the UAV position at time $t+1$ is updated via the uniformly accelerated motion model that
\begin{align}
 {{\mathbf{q}}_U}\left( {t + 1} \right) = {{\mathbf{q}}_U}\left( t \right) + {{\mathbf{v}}_U}\left( t \right)\Delta t + 0.5{{\mathbf{a}}_U}\left( t \right)\Delta {t^2},
\end{align}
where $\Delta t$ refers to the duration of a single time slot.

Since the electromagnetic reflection characteristics of RIS are highly dependent on the incident angle of electromagnetic waves, three attitude angles, including yaw ${\zeta _y}$, pitch ${\zeta _p}$ and roll ${\zeta _r}$, are modeled for the aerial RIS mounted on the UAV. Considering the indoor scenario adopted in this work, the three-dimensional rotation angles of the UAV are precisely controlled without external disturbances such as wind-induced jitter. The quadrotor UAV is considered in this paper, whose translational moving direction is decoupled from the heading direction of the nose. Therefore, a 3D rotation matrix is constructed from the above three attitude angles as 
\begin{align}
  {\mathbf{R}} = &{{\mathbf{R}}_{yaw}}{{\mathbf{R}}_{pitch}}{{\mathbf{R}}_{roll}}\nonumber\\
  = &\left[ {\begin{array}{*{20}{c}}
  {\cos {\zeta _y}}&{ - \sin {\zeta _y}}&0 \\ 
  {\sin {\zeta _y}}&{\cos {\zeta _y}}&0 \\ 
  0&0&1 
\end{array}} \right]\left[ {\begin{array}{*{20}{c}}
  {\cos {\zeta _p}}&0&{\sin {\zeta _p}} \\ 
  0&1&0 \\ 
  { - \sin {\zeta _p}}&0&{\cos {\zeta _p}} 
\end{array}} \right]\nonumber\\
&\times \left[ {\begin{array}{*{20}{c}}
  1&0&0 \\ 
  0&{\cos {\zeta _r}}&{ - \sin {\zeta _r}} \\ 
  0&{\sin {\zeta _r}}&{\cos {\zeta _r}} 
\end{array}} \right].
\end{align}

\begin{remark}\label{remark:angle caculation}
For an RIS arranged parallel to the $xOy$ plane, the corresponding normal vector, $x$-axis directional vector and $y$-axis directional vector are defined as ${{\mathbf{\overset{\lower0.5em\hbox{$\smash{\scriptscriptstyle\rightharpoonup}$}} {n} }}_{RIS}} = {\left[ {0,0, 1} \right]^T}$, ${{\mathbf{e}}_x} = {\left[ {1,0,0} \right]^T}$ and ${{\mathbf{e}}_y} = {\left[ {0,1,0} \right]^T}$. After the UAV undergoes three-axis attitude deflections characterized by ${\zeta _y},{\zeta _p},{\zeta _r}$, the update process of RIS vectors is obtained through coordinate rotation: ${\left( {{{\mathbf{n}}_{RIS}},{{\mathbf{e}}_x},{{\mathbf{e}}_y}} \right)_{t + 1}} = {{\mathbf{R}}_t}\left( {{{\mathbf{n}}_{RIS}},{{\mathbf{e}}_x},{{\mathbf{e}}_y}} \right)$.

Combining the established UAV motion and attitude models, the precise incident angle of electromagnetic waves impinging on RIS from transmitter $DU_k$ is calculated as
\begin{align}
{\theta _{k,in}} = \arccos \left( {\frac{{\left\langle {{{\mathbf{q}}_U} - {{\mathbf{q}}_k},{{{\mathbf{\overset{\lower0.5em\hbox{$\smash{\scriptscriptstyle\rightharpoonup}$}} {n} }}}_{RIS}}} \right\rangle }}{{\left| {{{\mathbf{q}}_U} - {{\mathbf{q}}_k}} \right|\left| {{{{\mathbf{\overset{\lower0.5em\hbox{$\smash{\scriptscriptstyle\rightharpoonup}$}} {n} }}}_{RIS}}} \right|}}} \right),
\end{align}
where ${{\mathbf{q}}_k}$ denotes the spatial coordinate of transmitter $DU_k$. Similarly, the spatial frequencies at RIS is calculated as
\begin{align}
  {u_{k,R}} = {\frac{{\left\langle {{{\mathbf{q}}_U} - {{\mathbf{q}}_k},{{\mathbf{e}}_x}} \right\rangle }}{{\left| {{{\mathbf{q}}_U} - {{\mathbf{q}}_k}} \right|\left| {{{\mathbf{e}}_x}} \right|}}} , {v_{k,R}} = {\frac{{\left\langle {{{\mathbf{q}}_U} - {{\mathbf{q}}_k},{{\mathbf{e}}_y}} \right\rangle }}{{\left| {{{\mathbf{q}}_U} - {{\mathbf{q}}_k}} \right|\left| {{{\mathbf{e}}_y}} \right|}}} .
\end{align}
It is worth noting that the calculated incident angle accounts for the three-dimensional rotational attitudes of the UAV, representing the true incidence angle of transmitted signals impinging on the RIS plane, and corresponds to AOA and AOD. This angular parameter is substituted into \eqref{eq:Rician} and \eqref{eq:practical r} to acquire the updated Rician channel as well as the practical reflection coefficient of RIS.
\end{remark}

\section{Problem Formulation}
In this work, we investigate UAV-mounted RIS assisted communications for multiple D2D pairs, where the practical RIS reflection coefficient varies with wave incident angles. Accordingly, we aim to maximize the sum achievable rate of all D2D pairs via jointly optimizing UAV trajectories, UAV attitudes, and preset RIS reflection coefficient. The optimization problem is formulated as follows:
\begin{align}\label{eq:problem}
  \mathop {\max }\limits_{\left\{ {{\mathbf{Q}}\left( t \right),{\mathbf{\Theta }}\left( t \right)} \right\}} &\frac{1}{T}\sum\limits_{t = 1}^T {\sum\limits_{k = 1}^K {{R_k}} }  \\
  s.t.&{{\mathbf{q}}_U}\left( t \right) \in {\Omega _U}\tag{\theequation a} \label{eq:f}\\
  {\text{  }}&\left| {{{\mathbf{v}}_U}\left( t \right)} \right| \leqslant {v_{\max }} \tag{\theequation b} \label{eq:a}\\
  &\left| {{{\mathbf{a}}_U}\left( t \right)} \right| \leqslant {a_{\max }} \hfill \tag{\theequation c} \label{eq:b}\\
  &\left| {{r_n}} \right| \leqslant 1,\angle {r_n} \in \left[ {0,2\pi } \right)  \tag{\theequation d} \label{eq:c}\\
  &{\zeta _y}\left( t \right),{\zeta _p}\left( t \right),{\zeta _r}\left( t \right) \in \left[ {0,2\pi } \right) \tag{\theequation e} \label{eq:d}\\
  &{\beta _k}\left( t \right) \in \left\{ {0,1} \right\}\tag{\theequation f}\label{eq:e}\\
  &\sum\nolimits_{t = 1}^T {{P_U}\left( t \right)}  \leqslant E_U^{\max }\tag{\theequation g}\label{eq:g}
\end{align}
where $T$ is the number of total time slots, ${\mathbf{Q}}\left( t \right) = \left\{ {{{\mathbf{q}}_U}\left( t \right),{\zeta _y}\left( t \right),{\zeta _p}\left( t \right),{\zeta _r}\left( t \right)} \right\}$ contains the UAV trajectories and UAV attitudes. \eqref{eq:f} constrains the UAV position, where $\Omega _U$ denotes the feasible flight region. \eqref{eq:a} and \eqref{eq:b} constrain the resultant velocity and acceleration, respectively. The amplitudes and phase shifts limitation is presented in \eqref{eq:c}. \eqref{eq:d} describes the full rotational range for three-axis UAV attitude adjustment at each time slot. \eqref{eq:e} is the binary indicator constrain for the $k$-th D2D pair. ${\beta _k}\left( t \right) = 1$ means the $k$-th D2D link is served at time slot $t$, while ${\beta _k}\left( t \right) = 0$ means this link remains inactive in the current time slot. Constraint (15d) limits the UAV’s total propulsion energy consumption to $E_U^{\max }$.

\section{Decision Transformer Optimization Framework}

To address the formulated joint optimization problem and improve the adaptability of the learned policy to different UAV-RIS deployment scenarios, we develop a generalizable Decision Transformer (DT) framework. The overall framework consists of four main components: MDP formulation, DRL-based expert algorithm selection, offline DT pre-training, and online fine-tuning for unseen scenarios.

\begin{figure}[t]
  \centering
  \includegraphics[scale=0.3]{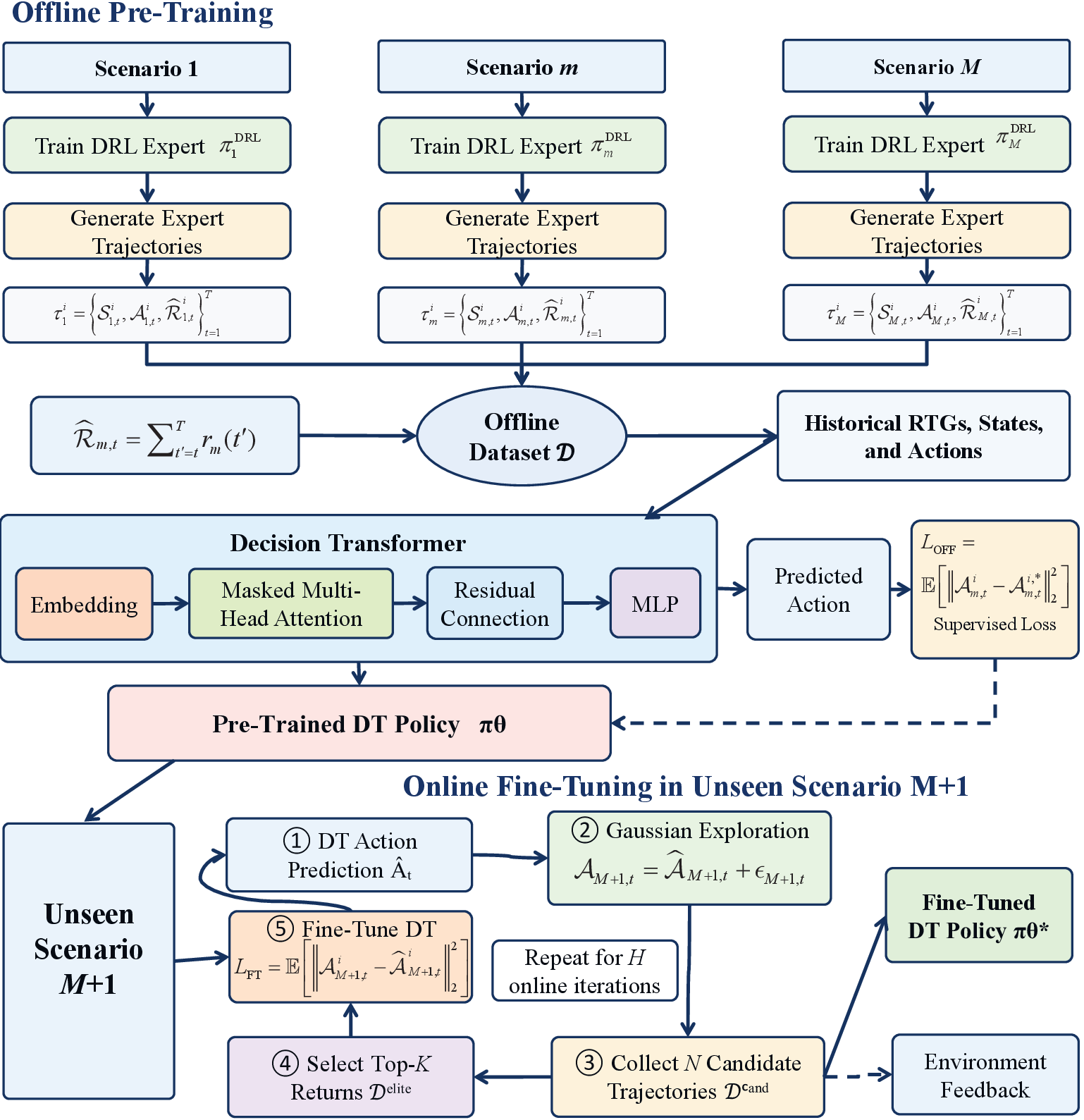}
  \caption{Decision Transformer Framework with Offline Pre-Training and Online Fine-Tuning}\label{ss}
\end{figure}

\subsection{MDP Formulation}
We first model the system as a MDP, which consists of a state space $\bm{\mathcal{S}}$, an action space $\bm{\mathcal{A}}$, a state transition probability $\bm{\mathcal{P}}$, and a reward function $\bm{\mathcal{R}}$, and denoted by the tuple $\left\langle \bm{{\mathcal{S},\mathcal{A},\mathcal{P},\mathcal{R}}} \right\rangle $, which are defined as follows. 

1) \textit{State:} The state space at timestep $t$ is defined by ${\mathcal{S}_t} = \left\{ {{{\mathbf{q}}_U}\left( t \right),{{\mathbf{v}}_U}\left( t \right),{\zeta _y}\left( t \right),{\zeta _p}\left( t \right),{\zeta _r}\left( t \right), {\mathbf{H}}\left( t \right), \mathbf{P}\left( t \right)} \right\}$, where ${\mathbf{H}}\left( t \right)\in {\mathbb{R}^{N \times K}}$ denotes the Rician fading matrix and ${\mathbf{P}}\left( t \right) \in {\mathbb{R}^{K \times K}}$ represents the D2D pair state in time $t$.

2) \textit{Action:} The action space is defined as: $\mathcal{A}_t=\{\mathbf{a}_{U}(t),\mathbf{\Theta}(t), \Delta {\zeta _y}\left( t \right), \Delta {\zeta _p}\left( t \right), \Delta {\zeta _r}\left( t \right)\}$, where $\Delta {\zeta_y}(t)$, $\Delta {\zeta_p}(t)$, and $\Delta {\zeta_r}(t)$ denote the incremental changes of 3D rotation angles.

3) \textit{Transition probability:} After taking action $\mathcal{A}_t$ in the current state ${\mathcal{S}_t}$, the system transits to the next state ${\mathcal{S}_{t+1}}$. The transition probability is denoted as $\mathcal{P}\left( {{\mathcal{S}_{t + 1}}\mid {\mathcal{S}_t},{\mathcal{A}_t}} \right)$, which represents the probability of moving from $\mathcal{S}_t$ to $\mathcal{S}_{t+1}$ after executing $\mathcal{A}_t$.

4) \textit{Reward:} At each time slot $t$, the reward is defined as $r(t)=\sum\nolimits_{k = 1}^K {{R_k}}(t) $. The reward return is defined as ${\mathcal{R}_t} = \sum\nolimits_{\tau  = t}^T {{\gamma ^{\tau  - t}}r(\tau )}$, where $\gamma$ is the discount factor. The objective of the MDP is to learn an optimal policy $\pi^*$ that maximizes the expected cumulative reward ${\pi ^*} = \arg {\max _\pi }{E_\pi }\left[ \mathcal{R}_1 \right]$.

\subsection{DRL-Based Expert Algorithm Selection} 

To construct a high-quality offline dataset, several representative DRL algorithms for continuous control are first trained and evaluated under the same simulation settings, including proximal policy optimization (PPO)\cite{PPO}, DDPG\cite{DDPG}, SAC\cite{SAC}, and twin delayed deep deterministic policy gradient (TD3)\cite{TD3}. Their convergence behavior, average sum-rate performance, and policy stability are compared, and the best-performing algorithm is selected as the expert. The detailed comparison is presented in Section V. This empirical selection procedure avoids assuming in advance that a particular DRL algorithm is optimal for the considered environment.

\subsection{Offline Pre-Training}

First, the selected DRL algorithm is independently trained in $M$ different UAV-RIS scenarios to obtain high-quality expert policies. The $i$-th trajectory generated by DRL policy $m$ can be represented as
\begin{align}
 \tau_m^i=\left\{\mathcal{S}_{m,1}^i,\mathcal{A}_{m,1}^i,\hat{\mathcal{R}}_{m,1}^i,\ldots,\mathcal{S}_{m,T}^i,\mathcal{A}_{m,T}^i,\hat{\mathcal{R}}_{m,T}^i\right\},
\end{align}
where $\hat{\mathcal{R}}_{m,t}$ denotes the return-to-go (RTG) from time slot $t$, which is calculated as
\begin{align}
 \hat{\mathcal{R}}_{m,t}=\sum\nolimits_{t'=t}^{T} r_m(t').
\end{align}
Multiple trajectories can be generated by each of the $ M$  expert policies and added to the dataset $\mathcal{D}$. During offline training, DT takes the historical states, actions and RTGs as input and predicts the action at the current time slot. The model parameters are optimized by minimizing the mean squared error between the predicted action $\mathcal{A}_{m,t}^{i,*}$ and the expert action $\mathcal{A}_{m,t}^i$, i.e.,
\begin{align}\label{eq:Loss-OFFLINE}
L_{\mathrm{OFF}}=\mathbb{E}_{\tau_m^i\sim\mathcal{D}}\left[\left\|\mathcal{A}_{m,t}^i-\mathcal{A}_{m,t}^{i,*}\right\|_2^2\right].
\end{align}

Through multi-scenario offline pre-training, DT learns the relationship between system states, target returns, and high-quality control actions, thereby obtaining an initialized policy with cross-scenario decision-making capability.

\subsection{Online Fine-Tuning}

When the UAV-RIS system encounters a new scenario $M+1$ that is not contained in the offline training dataset, the pre-trained DT is first directly deployed to generate a zero-shot control policy. Given the initial target return $\hat{\mathcal{R}}_{M+1,1}$ and the observed state $\mathcal{S}_{M+1,1}$, DT predicts the corresponding action $\mathcal{\hat A}_{M+1,1}$. After executing the action, the RTG is updated according to $\hat{\mathcal{R}}_{M+1,t+1}=\hat{\mathcal{R}}_{M+1,t}-r(t)$, and the updated RTG, system state, and historical actions are subsequently used to generate the next control action. Notably, to encourage exploration in the new scenario, Gaussian noise with standard deviation \(\sigma\) is added to the action predicted by DT. Thus, the action executed during online interaction is given by \(\mathcal A_{M+1,t}=\hat{\mathcal A}_{M+1,t}+\epsilon_{M+1,t}\), where \(\epsilon_{M+1,t}\sim\mathcal N(0,\sigma^2\mathbf I)\).

After completing multiple episodes, the newly collected samples are organized into a set of candidate trajectories, where the $i$-th trajectory of new scenario is represented as
\begin{align}
 \tau_{M+1}^{i}=&\left\{\mathcal S_{M+1,1}^{i},\mathcal A_{M+1,1}^{i},\hat{\mathcal R}_{M+1,1}^{i},\ldots,\right.\nonumber\\
 &\left.\mathcal S_{M+1,T}^{i},\mathcal A_{M+1,T}^{i},\hat{\mathcal R}_{M+1,T}^{i}\right\}.
\end{align}
Rather than using all collected trajectories for fine-tuning, the candidate trajectories are evaluated according to their cumulative returns. Only those with the top-\(K\) returns are selected to construct an elite trajectory set
\begin{align}
\mathcal D_{M+1}^{\mathrm{elite}}
=
\operatorname{TopK}_{\tau_{M+1}^{i}}
\left(
\sum\nolimits_{t=1}^{T_i} r_i(t)
\right).
\end{align}
The selected trajectories serve as high-quality references for fine-tuning the pre-trained DT, whereas trajectories with relatively poor performance are discarded. Specifically, the model is updated by minimizing the action prediction error over the elite trajectory set:
\begin{align}\label{eq:loss-finetuning}
L_{\mathrm{FT}}
=
\mathbb E_{\tau_{M+1}^{i}\sim
\mathcal D_{M+1}^{\mathrm{elite}}}
\left[
\left\|
\mathcal A_{M+1,t}^{i}
-
\hat{\mathcal A}_{M+1,t}^{i}
\right\|_2^2
\right],
\end{align}
where \(\mathcal A_{i,t}\) is the action recorded in a selected trajectory and \(\hat{\mathcal A}_{i,t}\) is the action predicted by DT. Through repeated trajectory collection, quality-based selection, and fine-tuning, the pre-trained model gradually adapts to the new UAV-RIS scenario while avoiding performance degradation caused by low-quality online trajectories and retaining the knowledge learned from the offline dataset.

The overall DT optimization framework based on DRL expert trajectories is summarized in \textbf{Algorithm \ref{algo:DT}}.
\begin{algorithm}[t]
\caption{DT Offline Pre-Training and Online Fine-Tuning}
\label{algo:DT}
\begin{algorithmic}[1]

\Require Offline scenarios $M$, online iterations $H$, candidate 
trajectories $N$, elite trajectories $K$, exploration standard deviation $\sigma$
\Ensure Fine-tuned DT policy $\pi_\theta$

\Statex \textbf{Stage 1: Offline pre-training}
\State Initialize offline dataset $\mathcal D\gets\varnothing$
\For{$m=1$ to $M$}
    \State Train DRL expert policy $\pi_m^{\mathrm{DRL}}$
    \State Generate multiple trajectories using $\pi_m^{\mathrm{DRL}}$
    \State Compute their RTGs and add them to $\mathcal D$
\EndFor
\State Pre-train DT on $\mathcal D$ using the loss in \eqref{eq:Loss-OFFLINE}

\Statex \textbf{Stage 2: Online fine-tuning}
\For{$h=1$ to $H$}
    \State Initialize candidate set 
    $\mathcal D_{M+1}^{\mathrm{cand}}\gets\varnothing$
    \For{$i=1$ to $N$}
        \State Generate $\tau_{M+1}^{i}$ using
        $\mathcal A_{i,t}=\hat{\mathcal A}_{i,t}+\epsilon_{i,t}$,
        $\epsilon_{i,t}\sim\mathcal N(\mathbf 0,\sigma^2\mathbf I)$
        \State Add $\tau_{M+1}^{i}$ to 
        $\mathcal D_{M+1}^{\mathrm{cand}}$
    \EndFor
    \State Select the top-$K$ trajectories to construct
    $\mathcal D_{M+1}^{\mathrm{elite}}$
    \State Fine-tune DT on $\mathcal D_{M+1}^{\mathrm{elite}}$
    using the loss in \eqref{eq:loss-finetuning}
\EndFor

\State \Return $\pi_\theta$

\end{algorithmic}
\end{algorithm}

\section{Numerical Results}

In the simulations, we consider a UAV-mounted RIS-assisted communication system deployed within a three-dimensional area of $40 \times 40 \times 8~\mathrm{m}^3$. The system consists of four D2D users located at $[15,5,1.0]$, $[20,8,1.5]$, $[30,20,0.1]$, and $[15,35,0.6]$, respectively, where all coordinates are measured in meters. RIS is configured as a $4\times4$ uniform planar array comprising 16 reflecting elements. The transmit power of each user is set to 1W. The Rician $\kappa$-factor and path-loss exponent set to \(\kappa=8\) and \(\alpha=2.4\), respectively. Unless otherwise specified, these parameter settings are adopted throughout the subsequent simulations.

\begin{figure}[t]
  \centering
  \includegraphics[width=6.5cm, height=5cm]{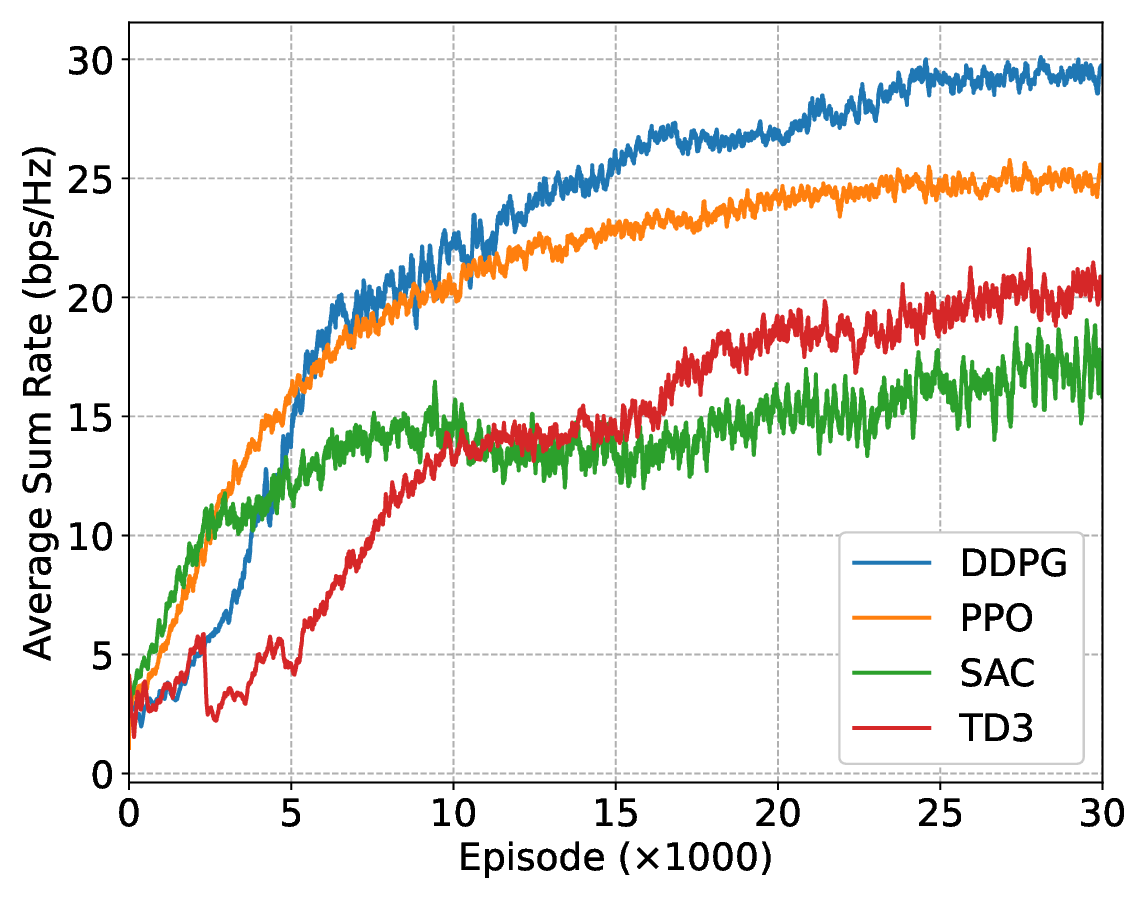}
  \caption{Convergence performance comparison of different DRL algorithms in the UAV mounted RIS-assisted communication environment.}\label{fig:episode reward}
\end{figure}

Fig. \ref{fig:episode reward} compares the training performance of four representative DRL algorithms, including DDPG, PPO, SAC, and TD3. In this figure, the initial position of UAV-RIS is set to [15, 20, 4.5]. We can observe that DDPG improves rapidly after an initial exploration stage, and gradually stabilizes after approximately 25,000 episodes, ultimately achieving an average sum rate of approximately 29.5 bps/Hz. This value is significantly higher than those obtained by PPO, TD3, and SAC, which converge to approximately 25.0, 20.5, and 17.0 bps/Hz, respectively.

The action space of the proposed environment consists of three-dimensional UAV motion control, UAV attitude adjustment, and the phase control of multiple RIS elements. Therefore, the resulting optimization problem is characterized by a continuous, high-dimensional, and strongly coupled action space. DDPG employs a deterministic policy gradient and can directly generate continuous control variables, thereby avoiding the dimensional expansion and loss of control accuracy caused by discretizing the action space. Moreover, its experience replay mechanism allows historical interaction data to be reused, improving sample efficiency, while the target networks help mitigate severe fluctuations during value-function estimation. These characteristics make DDPG well suited to the continuous joint optimization problem considered in this study. 

Based on the above comparison, the DDPG algorithm is selected as the expert policy for DT. Specifically, 8 training scenarios are considered, each characterized by a different initial horizontal position of the UAV-mounted RIS. The initial altitude is fixed at $z=4.5~\mathrm{m}$, while the corresponding $[x,y]$ coordinates are set to $[10,10]$, $[10,30]$, $[15,20]$, $[20,20]$, $[20,30]$, $[25,20]$, $[30,10]$, and $[30,30]$, respectively. For each scenario, the converged DDPG expert policy is employed to generate 500 trajectories without additional exploration noise. These trajectories are subsequently aggregated to form the offline training dataset for DT. During the fine-tuning stage, 80 trajectories are collected through online interactions with the environment, from which the top 50 trajectories are selected for policy fine-tuning.

\begin{figure}[t]
  \centering
  \subfloat[]{%
    \includegraphics[width=0.48\linewidth]{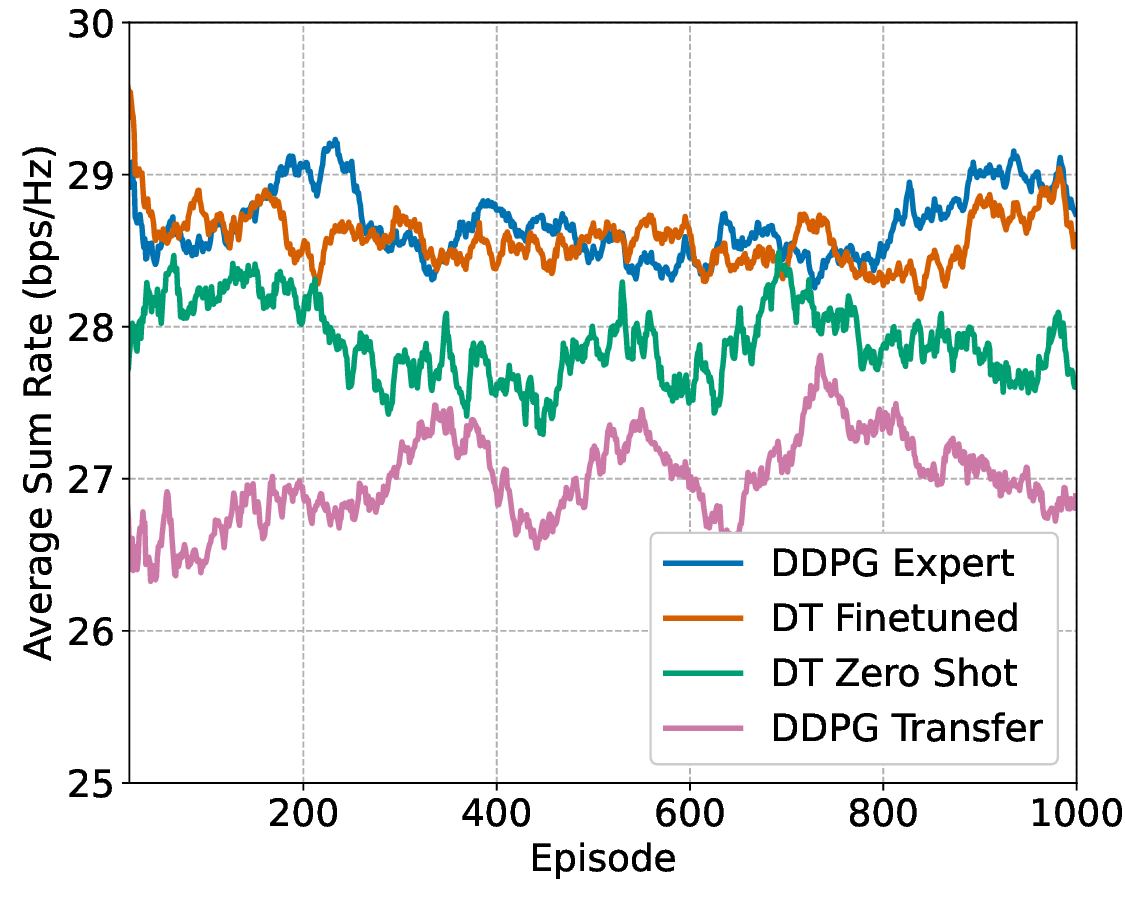}
    \label{fig:DT_verify_a}}
  \hfill
  \subfloat[]{%
    \includegraphics[width=0.48\linewidth]{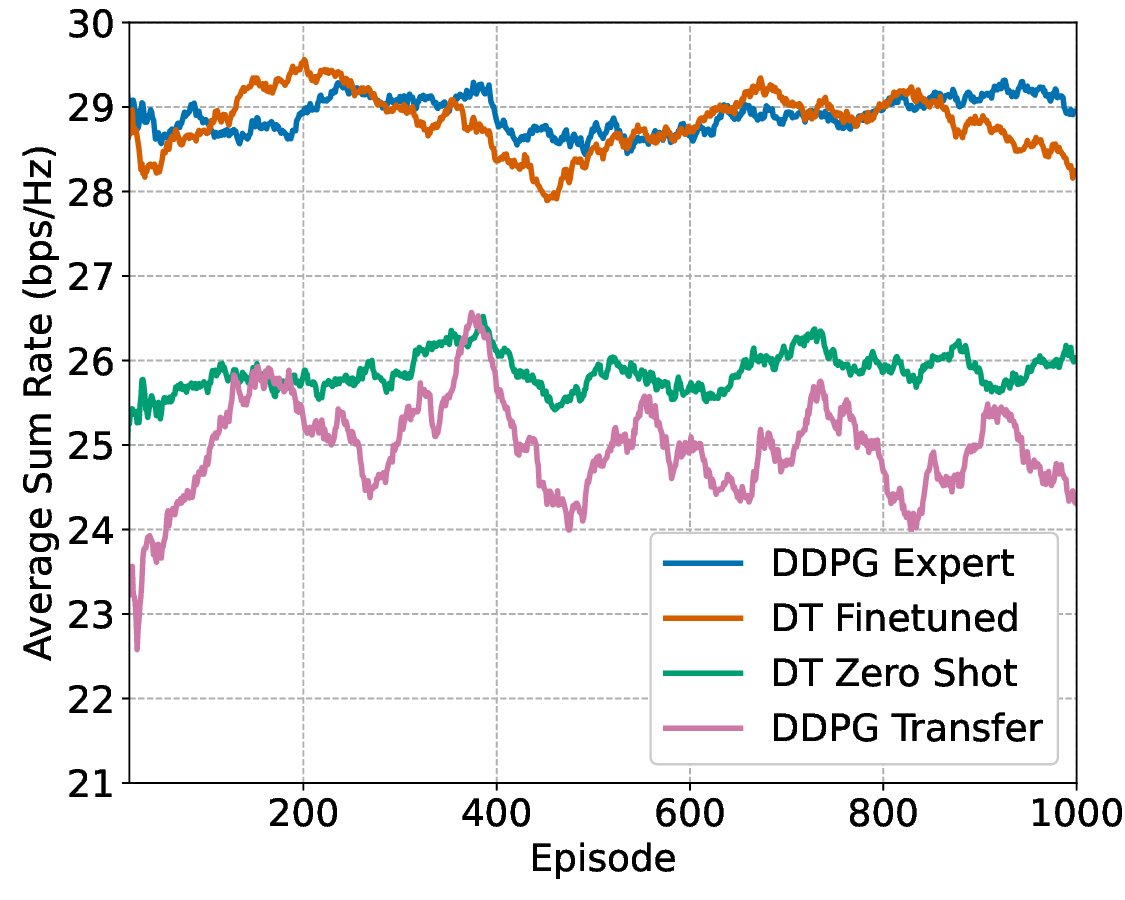}
    \label{fig:DT_verify_b}}
  \caption{Evaluation results of the DDPG expert, fine-tuned DT, zero-shot DT, and transferred DDPG policies in the unseen scenario. (a) Initial UAV-mounted RIS position at [15,15,4.5]. (b) Initial UAV-mounted RIS position at [25,25,4.5].}
  \label{fig:DT_verify}
\end{figure}

Fig. \ref{fig:DT_verify} compares the evaluation performance of different policies in the unseen scenario with the initial UAV-mounted RIS position set to $[15,15,4.5]$ in Fig. \ref{fig:DT_verify_a} and $[25,25,4.5]$ in Fig. \ref{fig:DT_verify_b}. This scenario is excluded from the offline dataset used to train DT and is therefore employed to evaluate policy generalization. The horizontal axis represents the evaluation episodes rather than the training process. 

As shown in Fig. \ref{fig:DT_verify}, the fine-tuned DT achieves nearly the same performance as the scenario-specific DDPG expert. The zero-shot DT also obtains a high average sum rate and consistently outperforms the DDPG policies transferred from neighboring scenarios. These results demonstrate that multi-scenario offline training enables DT to generalize effectively to an unseen UAV–RIS initial position, while fine-tuning further improves its performance to the expert level. In addition, we can also observe that the zero-shot DT and the transferred DDPG exhibit larger fluctuations because they have not been optimized directly for the target scene and are therefore more sensitive to distribution shifts in channel states, UAV trajectories, and D2D pairings.


\section{Conclusions}
This paper studied a UAV-mounted RIS-assisted D2D communication system with stochastic link activation, considering UAV motion, three-dimensional attitude, time-varying Rician LoS angles, and incident-angle-dependent RIS responses. A joint optimization problem was formulated to maximize the average sum rate through UAV trajectory, attitude, and RIS phase control. Among several DRL methods, DDPG generated expert trajectories for multiple scenarios. A Decision Transformer was pre-trained on this offline dataset and adapted to unseen scenarios through zero-shot deployment and online fine-tuning. Results showed that zero-shot DT outperformed direct DDPG transfer, while fine-tuned DT approached scenario-specific DDPG performance with fewer interactions, demonstrating the potential of expert-data-driven sequence modeling for generalizable and efficient UAV-RIS control.

%

\vspace{12pt}


\begin{thebibliography}{10}


%

%
%
%



\bibitem{RIS} M. Ahmed et al., “A comprehensive survey of artificial intelligence advances in reconfigurable intelligent surfaces-assisted wireless networks,” \emph{Engineering Applications of Artificial Intelligence}, vol. 176, Art. no. 114762, Jul. 2026.

\bibitem{EM1} B. Xu, T. Zhou, F. Gao, T. Xu, and H. Hu, “RIS-aided MIMO communications: Angle-dependent amplitude-phase response model and capacity analysis,” \emph{IEEE Wireless Commun. Lett.}, vol. 14, no. 2, pp. 290–294, Feb. 2025.

\bibitem{EM2} Y. Chen, Y. Guo, and H. Zhang, “Angle-sensitive effect of reconfigurable intelligent surface,” \emph{IEEE Trans. Wireless Commun.}, vol. 24, no. 11, pp. 9556–9568, Nov. 2025.

\bibitem{1-1} H. Zhao, W. Sun, Y. Ni, W. Xia, G. Gui, and C. Zhu, “Deep deterministic policy gradient-based rate maximization for RIS-UAV-assisted vehicular communication networks,” \emph{IEEE Trans. Intell. Transp. Syst.}, vol. 25, no. 11, pp. 15732–15744, Nov. 2024.

\bibitem{1-2} W. Chen, Y. Zou, J. Zhu, and L. Zhai, “Joint trajectory design and phase shift optimization for multi-RIS-assisted UAV relay network using deep reinforcement learning,” \emph{IEEE Internet Things J.}, vol. 12, no. 8, pp. 9759–9774, Apr. 2025.
    
\bibitem{3-1} M. M. Salim, K. M. Rabie, and A. H. Muqaibel, “Robust energy-efficient DRL-based optimization in UAV-mounted RIS systems with jitter,” \emph{IEEE Commun. Lett.}, vol. 29, no. 12, pp. 2780–2784, Dec. 2025.

\bibitem{3-2} C. Liu, W. Mei, and Z. Chen, “Joint 3D orientation and location optimization for UAV-mounted intelligent reflecting surface,” in \emph{Proc. IEEE Global Commun. Conf. (GLOBECOM)}, Cape Town, South Africa, Dec. 2024, pp. 2725–2730.

\bibitem{SAC-review} X. Zhou, L. Huang, T. Ye, and W. Sun, “Computation bits maximization in UAV-assisted MEC networks with fairness constraint,” \emph{IEEE Internet Things J.}, vol. 9, no. 21, pp. 20997–21009, Nov. 2022.

\bibitem{zhangjie} J. Zhang et al., “Decision transformers for wireless communications: A newparadigm of resource management,” \emph{IEEE Wireless Commun.}, vol. 32, no. 2, pp. 180–186, Apr. 2025.

\bibitem{DT} L. Chen et al., “Decision transformer: Reinforcement learning via sequence modeling,” in \emph{Proc. Int. Conf. Neural Inf. Process.}, 2021, vol. 34, pp. 15084–15097.

\bibitem{Unified} Y. Zhang et al., “A unified deterministic channel model for multi-type RIS with reflective, transmissive, and polarization operations,” \emph{IEEE Trans. Veh. Technol.}, vol. 75, no. 2, pp. 2821–2833, Feb. 2026.

\bibitem{PPO} J. Schulman, F. Wolski, P. Dhariwal, A. Radford, and O. Klimov, “Proximal policy optimization algorithms,” arXiv preprint arXiv:1707.06347, 2017.

\bibitem{DDPG} T. P. Lillicrap et al., “Continuous control with deep reinforcement learning,” in \emph{Proc. Int. Conf. Learn. Represent. (ICLR)}, 2016.

\bibitem{SAC} T. Haarnoja, A. Zhou, P. Abbeel, and S. Levine, “Soft actor-critic: Off-policy maximum entropy deep reinforcement learning with a stochastic actor,” in \emph{Proc. 35th Int. Conf. Mach. Learn. (ICML)}, vol. 80, pp. 1861–1870, 2018.

\bibitem{TD3} Y. Hou, H. Hong, Z. Sun, D. Xu, and Z. Zeng, “The control method of twin delayed deep deterministic policy gradient with rebirth mechanism to multi-DOF manipulator,” \emph{Electronics}, vol. 10, no. 7, Art. no. 870, Apr. 2021.

\end{thebibliography}
\end{document}